\documentclass[a4paper,11pt,twocolumn,twoside]{article}
\usepackage[dvips]{graphicx}
\usepackage{sepln_en}
\usepackage{fullname}
\usepackage[utf8]{inputenc}
\usepackage{tabularx} 
\usepackage{graphicx}
\usepackage{booktabs}
\usepackage[english]{babel}
\usepackage{url}
\usepackage{pdflscape}
\usepackage{changepage}
\input epsf

\title{Overview of ADoBo at IberLEF 2025: Automatic Detection of Anglicisms in Spanish}

\author{
\textbf{Elena Álvarez-Mellado$^1$}, 
\textbf{Jordi Porta-Zamorano$^2$}, \\ 
\textbf{Constantine Lignos$^3$},  
\textbf{Julio Gonzalo$^1$} \\
$^1$NLP \& IR group, UNED, Madrid, Spain\\
$^2$Laboratorio de Lingüística Informática, UAM, Madrid, Spain\\
$^3$Michtom School of Computer Science, Brandeis University, Massachusetts, USA\\
\{elena.alvarez, julio\}@lsi.uned.es, jordi.porta@uam.es, lignos@brandeis.edu\\
}

\seplntranstitle{Resumen de ADoBo 2025: detección automática de anglicismos en español}

\seplnclave{Préstamo léxico, anglicismos, detección automática de préstamos.}

\seplnresumen{En este artículo presentamos los resultados de ADoBo 2025, la tarea compartida de IberLEF 2025 sobre detección automática de anglicismos en castellano. La tarea consistió en identificar anglicismos contenidos en una colección de frases en castellano de estilo periodístico. Cinco equipos participaron en la fase de test y propusieron sistemas de diversa naturaleza (LLMs, \textit{deep learning}, sistemas basados en reglas, sistemas basados en Transformers) con resultados que oscilan entre los 0.17 y los 0.99 puntos de valor F1, lo que ilustra la variabilidad de resultados que distintos sistemas pueden obtener para esta tarea.
}

\seplnkey{Automatic detection of borrowings, loanword detection, linguistic borrowing, anglicisms.}

\seplnabstract{
This paper summarizes the main findings of ADoBo 2025, the shared task on anglicism identification in Spanish proposed in the context of IberLEF 2025. Participants of ADoBo 2025 were asked to detect English lexical borrowings (or anglicisms) from a collection of Spanish journalistic texts. Five teams submitted their solutions for the test phase. Proposed systems included LLMs, deep learning models, Transformer-based models and rule-based systems. The results range from F1 scores of 0.17 to 0.99, which showcases the variability in performance different systems can have for this task.}

\firstpageno{1}

\begin{document}

% la siguiente instrucción sólo se debe usar si el abstract sobrescribe el texto
% la longitud variará según se necesite

\setlength\titlebox{16cm} % se aumenta el tamaño del espacio reservado para datos de título

\label{firstpage} \maketitle

\section{Introduction}

Linguistic borrowing is the process of reproducing in one language elements and patterns that come from another language \cite{haugen1950analysis}. Linguistic borrowing therefore involves the exchange between two languages and has been widely studied within the field of contact linguistics \cite{weinreich1963languages}.
Lexical borrowing in particular is the process of importing words from one language into another \cite{poplack1988social,onysko2007anglicisms}.
Lexical borrowing is a phenomenon that occurs in all languages and is a prolific source of new words and meanings \cite{gerding2014anglicism}.

In recent decades, English in particular has produced numerous lexical borrowings (often called \textit{anglicisms}) in many European languages \cite{furiassi2012anglicization}. Previous work estimated that a reader of French newspapers encounters a new lexical borrowing every 1,000 words  \cite{chesley_paula_predicting_2010}, English borrowings outnumbering all other borrowings combined \cite{chesley2010lexical}. In Chilean newspapers, lexical borrowings account for approximately 30\% of neologisms, 80\% of those corresponding to anglicisms \cite{gerding2014anglicism}. In European Spanish, it was estimated that anglicisms could account for 2\% of the vocabulary used in Spanish newspaper \textit{El País} in 1991 \cite{gorlach_felix}, a number that is likely to be higher today. As a result, the usage of lexical borrowings in Spanish (and particularly anglicisms) has attracted lots of attention, both in linguistic studies and among the general public. 

The ADoBo shared task series proposes to work on the task of automatically identifying lexical borrowings from text. After a first shared task on 2021 \cite{mellado2021overview}, for this second edition on 2025 we propose a shared task on specifically detecting anglicisms from Spanish text. In this paper we describe the shared task held at IberLEF 2025 \cite{iberlef2025overview}: we introduce the systems that participated in it, and share the results obtained during the competition.

\section{Related work}

The task of extracting unassimilated lexical borrowings is a more challenging undertaking than it might appear to be at first. To begin with, lexical borrowings can be either single or multitoken expressions (e.g., \textit{prime time}, \textit{tie break} or \textit{machine learning}). Second, linguistic assimilation is a diachronic process and, as a result, what constitutes an unassimilated borrowing is not clear-cut. For example, words like \textit{bar} or \textit{club} were unassimilated lexical borrowings in Spanish at some point in the past, but have been around for so long in the Spanish language that the process of phonological and morphological adaptation is now complete and they cannot be considered unassimilated borrowings anymore. 

All these subtleties make the annotation of lexical borrowings non-trivial. Consequently, in prior work on anglicism extraction from Spanish text, plain dictionary lookup produced very limited results with F1 scores of 47 \cite{serigos2017applying} and 26 \cite{alvarez2020lazaro}. In fact, whether a given expression is a borrowing or not cannot always be determined by plain dictionary lookup; after all, an expression such as \textit{social media} is an anglicism in Spanish, even when both \textit{social} and \textit{media} also happen to be Spanish words that are registered in regular dictionaries. This justifies the need for a more NLP-heavy approach to the task.

%The automatic detection of borrowings has not been a major topic in NLP.
%Most of the work on automatic and semiautomatic approaches to borrowing detection has come from the realm of corpus linguistics, where the manual retrieval of borrowings of interest has traditionally been the norm.
%More recently some works have proposed different techniques to automatize the retrieval of borrowing candidates, initially through applying a series of heuristics to semiautomatically filter anglicism candidates based on pattern matching and character probability \cite{furiassi2012anglicization,andersen2012semi,serigos2017applying}, and more recently based on machine learning techniques \cite{losnegaard2012data}.

In the area of NLP, different automatic approaches to borrowing detection have been proposed. 
Lexical borrowing identification has proven to be relevant in lexicographic work as well as a pre-preprocessing step for NLP downstream tasks, such as parsing \cite{alex2008automatic} and text-to-speech synthesis \cite{leidig2014automatic}.
The identification of borrowings has also been used as bootstrapping technique in machine translation to enlarge the available vocabulary when working on very low-resourced languages that heavily borrow from high-resourced languages \cite{tsvetkov_constraint-based_2015,tsvetkov_cross-lingual_2016,mi2020loanword,mi2025multi}.

Several projects have approached the task of extracting lexical borrowings in various languages, such as German \cite{alex2008automatic,garley-hockenmaier-2012-beefmoves,leidig2014automatic}, Italian \cite{furiassi2007retrieval}, French \cite{alex2008automatic,chesley2010lexical}, Finnish \cite{mansikkaniemi_unsupervised_2012}, and Norwegian \cite{andersen2012semi,losnegaard2012data}, some of them with a particular focus on anglicism extraction, while others have taken a more language-agnostic approach to the problem \cite{nath_generalized_2022}.

Concretely for the task of retrieving anglicisms from Spanish text, various approaches have been proposed: lexicon lookup systems enriched with character n-gram probability \cite{serigos2017applying,serigos2017using}; semiautomatically filtering anglicism candidates based on lexicon lookup and pattern-matching \cite{moreno_fernandez_configuracion_2018}; a CRF model with handcrafted features \cite{alvarez2020lazaro}; a CRF model with data augmentation \cite{jiang2021bert4ever}; and several deep learning models and Transformer-based models \cite{de2021futility,alvarez-mellado-lignos-2022-detecting}.
In addition, a previous shared task held at IberLEF 2023 explored the task of retrieving Spanish lexical borrowings from a corpus of Guarani rich in codeswitches
\cite{chiruzzo_overview_2023}.

% \usepackage{booktabs}
% \usepackage[normalem]{ulem}
\begin{table}[]
\small
\resizebox{\textwidth}{!}{\begin{tabular}{@{}lrrr@{}}
\toprule
\textbf{Model}          & \multicolumn{1}{c}{\textbf{Precision}} & \multicolumn{1}{c}{\textbf{Recall}} & \multicolumn{1}{c}{\textbf{F1}} \\ \midrule
CRF                     & 62.21                                  & 6.50                                & 11.77                           \\
BETO                    & 76.05                                  & 23.55                               & 35.96                           \\
mBERT                   & 84.01                                  & \underline{23.80}                         & 37.09                           \\
BiLSTM-CRF A  & 82.74                                  & 23.55                               & 36.66                           \\
BiLSTM-CRF B & \underline{85.15}                            & 23.75                               & \underline{37.14}                     \\
8B-Llama3               & \textbf{90.96}                         & \textbf{36.37}                      & \textbf{51.96}                  \\ \bottomrule
\end{tabular}}
\caption{Precision, recall and F1 scores over spans on ADoBo 2025 task test set obtained by the 6 models proposed as baselines. Best result in bold, second best result underlined.}
\label{tab:baselines}
\end{table}
\section{Task description}
The proposed task for the 2025 edition of ADoBo consisted in identifying anglicisms (unassimilated lexical borrowings from the English language, such as \textit{running}, \textit{smartwatch}, \textit{influencer}, \textit{holding}, \textit{look}, \textit{hype}, \textit{prime time} and \textit{lawfare}) in a test set made of sentences in Spanish from the journalistic domain.
Participants were given an unannotated version of the test set and they were expected to return a version of the test set annotated by their system.

\begin{table*}[t]
\footnotesize
\centering
\resizebox{\textwidth}{!}{\begin{tabular}{lrrrrrrl}
\toprule
\multicolumn{1}{c}{\textbf{Team submission}} 
& \multicolumn{1}{c}{\textbf{Precision}}	
& \multicolumn{1}{c}{\textbf{Recall}}	
& \multicolumn{1}{c}{\textbf{F1}}
& \multicolumn{1}{c}{\textbf{TP}}	
& \multicolumn{1}{c}{\textbf{FP}}	
& \multicolumn{1}{c}{\textbf{FN}} 
& \multicolumn{1}{c}{\textbf{Reference}}
\\
\midrule
297754-qilex &
\textbf{98.84} & 
\textbf{98.74} & 
\textbf{98.79} &
2050 &
24 &
26 &
\namecite{qilex}
\\
297656-shentzu &
96.68 & 
95.47 & 
96.07 &
1982 &
68 &
94 &
\namecite{shentzu}
\\
298600-mheredia &
92.60 &
94.07 &
93.33 &
1953 &
156 &
123 &
\namecite{mheredia}
\\
292548-trockti & 
96.20 &
87.86 &
91.84   &
1824 &
72 &
252 &
\namecite{trockti}
\\
\bottomrule
\end{tabular}}
\caption{Leaderboard results on the test set. For each team, precision, recall and F1 score are provided, along with the reference number of borrowings, the true positives (TP), false positives (FP) and false negatives (FN).}
\label{tab:leaderboard}
\end{table*}

\subsection{Dataset}
We did not provide participants with any training set whatsoever. 
Participants were encouraged to use any resource at their disposal to train their systems (lexicons, rules, available corpora, the dataset from the 2021 edition of ADoBo, etc.).
We did provide participants with a development set, so they could evaluate their systems and refine them. 
The development set we released was a version of the development set from \namecite{alvarez-mellado-lignos-2022-detecting} (which was the same development set used in the 2021 edition of ADoBo), filtered so it would only include sentences that contained anglicisms (and no lexical borrowings from other languages).

The test set for the task was BLAS (Benchmark for Loanwords and Anglicisms in Spanish) \cite{alvarez2025}. 
BLAS is a small collection of linguistically-motivated sentences made by hand that aim to exhaustively cover the linguistic variability (in terms of shape, sentence position, casing, punctuation, etc.) in which an anglicism may appear. 
BLAS consists of 1,836 annotated sentences in Spanish (37,344 tokens), which contain 2,076 spans labeled as anglicisms. 
Every sentence in BLAS contains at least one span labeled as anglicism.
The anglicism spans contained in BLAS appear in different settings (the same spans will appear in different sentence positions, with different casing configurations, with and without quotation marks, etc.), so that we can assess how good a model is at retrieving anglicisms in certain contexts or identify systematic errors in performance.

\subsection{Evaluation}
The evaluation for the shared task was span based. This means that the expected output for each sentence in the test set was a list of spans of text, not BIO-encoded token annotations (unlike the 2021 edition). The scoring script expected a CSV file with semicolon separated values:
\begin{center}   
\texttt{
sentence;span1;span2;span3;etc.
}
\end{center}

In terms of metrics, standard precision, recall and F1 score over strict spans were used as evaluation metrics. 
This means that for a span to be considered to correct it had to match the span in the gold standard (no partial matches were considered). 

Our scoring script made the following assumptions:
\begin{itemize}
    \item[-] Casing of the output span is disregarded (\textit{SMARTWATCH} and \textit{smartwatch} will both match, regardless of which way it was written in the input sentence).
    \item[-] Trailing quotation marks in the output span were disregarded (\textit{“smartwatch”} and \textit{smartwatch} will both match, regardless of which way it was written in the input sentence).
    \item[-] If the same span appeared twice in the sentence, it sufficed for it to appear once in the output to be considered a match.
\end{itemize}

These assumptions were made in order to accommodate the participation of LLM-based solutions to the task.

\subsection{Baselines}
\label{subsec:baselines}
We proposed six baselines for the task of retrieving anglicisms from the test set: five supervised models already fine-tuned for the task of retrieving unassimilated lexical borrowings from Spanish text \cite{alvarez-mellado-lignos-2022-detecting} and one LLM on a few-shot approach (8B-Llama3) \footnote{\url{https://github.com/meta-llama/llama-models/blob/main/models/llama3/MODEL_CARD.md}} \cite{grattafiori2024llama3herdmodels,llama3modelcard}.

The five supervised models are a CRF \cite{alvarez2020lazaro}, fine-tuned BETO \cite{CaneteCFP2020}, fine-tuned mBERT \cite{devlin-etal-2019-bert} and two BiLSTM-CRFs: one of them fed with a combination of contextual embeddings based on BERT and BETO along with character and BPE embeddings (BiLSTM-CRF A), while the other was fed with contextual embeddings pretrained for the task of codeswitch identification on the English-Spanish section of the LinCE dataset \cite{aguilar-etal-2020-lince} (BiLSTM-CRF B). 
All of these supervised models were trained on the COALAS dataset from \namecite{alvarez-mellado-lignos-2022-detecting}.

Table \ref{tab:baselines} displays results for our six proposed baselines. 
8B-Llama3 outperforms all models across all three metrics (F1=51.96). Still, recall scores remain mediocre for all models, with a maximum of R=36.37 for 8B-Llama3 and a minimum of only R=6.50 for the CRF. 
Consequently, overall F1 scores remain modest across all models.
These baselines results showcase that there is ample room for improvement in this task and that the problem of lexical borrowing identification is far from being solved.

\section{Participating systems}
The shared task was held in Codabench\footnote{\url{https://www.codabench.org/competitions/7284/}}. 
Fourteen teams registered to participate in the competition. 
Overall we received 38 submissions from 6 different teams: 11 of those 38 submissions were on the development set and 27 of them corresponded to the test set, of which 4 were submitted to the leaderboard. 
Table \ref{tab:leaderboard} reports full results (precision, recall and F1 score) for the top four submissions that were submitted to the leaderboard.
Table \ref{tab:errors} displays an analysis of the errors made by the top performing system.
Tables \ref{tab:dev} and \ref{tab:test} report F1 score for all submissions made to the development set and the test set respectively.

Five out of the six participating teams submitted their outputs during the test phase. 
We now briefly present the five systems that were submitted by those five teams. 
We refer the reader to each of the system description papers for further details.

\subsection{Qilex team \cite{qilex}}
\namecite{qilex} explored several OpenAI LLMs for the task of retrieving anglicisms from Spanish text: 4.1, 4.1 mini, 4.1 nano, o4-mini and o3. 
They also throughly experimented with different prompting techniques (extended guidelines, self-refinement, chain-of-thought, in-context learning).
Their scores ranged from 12 to 99 of F1 score.
The best result was obtained by o3 model when prompting included explicit guidelines along with reminders. This combination produced the highest score overall in the shared task test set: 99 of F1 score.

\subsection{Shentzu team \cite{shentzu}}
\namecite{shentzu} experimented with a rule-based approach to the task. 
Their pipeline relied on a semi-automatically collected gazetteer of 37,000 lexical borrowing candidates extracted from a corpus of Spanish news of 6,600M tokens compiled by leveraging typographic conventions used in journalistic writing (quotation marks, italics) that was partially revised by a human. 
The gazetteer was the backbone of the rule-based pipeline, and optionally added an NER pre-processing step to ignore named entities and an already existing deep learning model.  

They conducted several experiments and combinations, the best solution yielding an F1 score 96 (which ranked \#2 on the leaderboard), obtained by a pure lexicon-based solution with rules that take into account some contextual features.

\subsection{Mheredia team \cite{mheredia}}
\namecite{mheredia} used instruction-tuned 70B-Llama 3.3 model to identify anglicism spans, along with several Transformer-based models.
They conducted experiments with zero-shot and few-shot prompting strategies, as well as the potential integration of auxiliary modules to improve performance. 

Regarding the models, they conducted experiments with Transformer-based models such as ModernBERT, BETO, IXABERT, XLM-RoBERTa large, and mDeBERTa v3, framing the task as a sequence labeling problem. As a decoder-only model, Llama 3.3 70B was used, with instructions specifying that the output must follow a predefined JSON structure. Using this model, prompts with various instructions based on the annotation guidelines were tested, both in zero-shot and 5-shot settings, and formulated in both English and Spanish. 

To improve the model’s precision, several modules were implemented, including a preliminary binary classifier that filters out texts not containing anglicisms. A list-based module was also used to identify and exclude named entities that are often mistakenly classified as anglicisms. Ultimately, an instruction was added to the prompt to help distinguish these entities from actual anglicisms. 

%A shift is observed in the distribution of anglicisms between the development set used and the test data. Additionally, they include a discussion on how to approach the task in terms of coverage and precision under different distributions using different settings.

Their best result was an F1 score of 93, which ranked \#3 on the leaderboard and was obtained by the instruction-based model with 5-shot prompting, without any of the classification or named entity recognition modules.

\subsection{Trockti team \cite{trockti}} 
\namecite{trockti} presented the use of Transformer-based models such as BERT and XLM-R to address the task of anglicism identification as a token classification or sequence labeling problem, following the BIO scheme. Based on an analysis of the model’s errors, they added a post-processing module that searches the \textit{Real Academia Española} dictionary \cite{dle} and uses spaCy \cite{honnibal2017spacy} to identify false positives caused by foreign named entities. They also performed an analysis of the distribution of the anglicisms in the development and test datasets and pointed several future extensions to their work.
Their best system obtained an F1 score of 91.84, which ranked \#4 on the leaderboard. 

\subsection{Hammond team \cite{hammond}}
\namecite{hammond} experimented with two systems for automatic detection of anglicisms in Spanish: one based on a logistic regression model and another on a feedforward neural network (FFNN). Both systems used a set of handcrafted binary features to classify words in a text as anglicisms or not. These features included orthotypographic and morphological information, lexicon lookup, character and bigram patterns, part-of-speech tags, and potential morphological stems as words in Spanish and English. On top of the output of these systems, several heuristic modules were applied to identify multiword units.

Their best result obtained an F1 score of 75, although their results were not submitted to the official leaderboard.

\section{Discussion}
\subsection{Results}
All of the outputs that were submitted to the leaderboard score above 90 (see Table \ref{tab:leaderboard}), and clearly outperform the best results obtained by our baselines, which were produced by 8B-Llama3 on few shot approach (see Table \ref{tab:baselines}). 
In fact, 25 out of the 27 submissions made during the test phase surpassed our baseline results (see Table \ref{tab:test}). 

The best overall F1 score results were obtained by the system submitted by qilex, based on o3 model when prompting was enriched with guidelines and reminders \cite{qilex} (F1=98.79). Qilex system ranked first both in terms of precision and recall, and also obtained the highest number of retrieved anglicisms and the lowest number of false positive and false negatives.
While these were the best results obtained by qilex, their paper also reports the impact that different models or different prompting methodologies can have on results: for instance, the very same OpenAI's o3 model scored an overall F1 of only 45 when no guidelines were added to the prompt.
On the other hand,  the very same guideline-enriched prompting methodology yielded a result of only 24 when they were applied to 4.1 Nano model. Quilex's thorough experiments showcase the high variability of results that LLMs can produce.

Qilex results were followed by shentzu's rule-based system \cite{shentzu} (F1 score=96.07). 
Shentzu's  results show that rule-based systems can succeed at this task while being less computing intensive than other solutions.
However, their solution relies on having an extensive pre-compiled lexicon of already-existing borrowings, which makes their solution less reliable to retrieve novel (i.e. previously unregistered) borrowings.  
Shentzu's  results are followed by the LLM-based system proposed by \namecite{mheredia} using 70B-Llama3.3 model (F1=93.33) and the Transformer-based systems by trockti \cite{trockti} (F1=91.84). 
Finally, although hammond did not submit their results to the leaderboard, their results illustrate that simple methods such as logistic regression and a feed-forward neural net fed with linguistic features can obtain competitive results for this task \cite{hammond}.

\subsection{Error analysis}
In Table~\ref{tab:errors} we present the instances in which the top-performing system of the shared task submitted by \namecite{qilex} produced incorrect predictions. The test set was deliberately constructed using orthotypographic variants of identical sentences to prevent systems from relying on orthotypography-specific cues and to identify systematic errors in performance. 

We classify the errors following the extended error typology proposed in \namecite{alvarez2025} inspired by the MUC error typology from \namecite{chinchor_muc-5_1993}, which considers the following error types:
\begin{itemize}
    \item[-] Missing:
  A span in the gold standard is not found in the prediction. 
\item[-] Spurious:
  A span in the prediction is not found in the gold standard.
\item[-] Type:
  The span in the prediction has a different label than the span in the gold standard.
\item[-] Overlap missing:
  The predicted span partially matches the gold standard span, but at least 1 token is missing.
\item[-] Overlap spurious:
  The predicted span partially matches the gold standard span, but at least 1 token is spurious.
\item[-] Split:
  One multiword span from the gold standard was retrieved as two adjacent shorter spans.
  \item[-] Fused:
  Two adjacent spans in the gold standard were retrieved as one long span.
\item[-] Missegmented:
  Two adjacent spans in the gold standard were retrieved as two adjacent spans, but the boundary between them was wrong.
\end{itemize}

As observed, the number of failed predictions in Table~\ref{tab:errors} is relatively small, with the corresponding instances grouped into ten clusters. The most frequent type of error is the system fully missing a span, followed by errors caused by overlapping spans (a span was partially retrieved, but a token was missed), split spans and finally fused spans.

In terms of which spans tended to cause the errors, the system seems to consistently fail when retrieving anglicisms that include ambiguous words that can exist both in English as well as Spanish, such as \textit{total red}, \textit{total black}, \textit{casual looks}, \textit{fatal error}, \textit{global director}, \textit{pie} or \textit{natural time}. 
In some of these examples, casing and quotation marks seems to help the system identify the span, but in others the model fails to capture them, regardless of orthotypographic variation, or only predicts partial elements or treats them as disjoint units. On the other hand, some adjacent spans that should be retrieved separately are retrieved as a single span, as in \textit{marketing} and \textit{online}. 

\subsection{Limitations and future work}
The near-perfect F1 scores that participants achieved on the ADoBo 2025 shared task (with the best-performing system scoring 99 of F1) raises an obvious question: is anglicism retrieval in Spanish effectively solved by the current generation of LLMs? Is there something left to be done for this task?

An important fact to bear in mind when analyzing results on this shared task is that BLAS, the test set used, is a dataset designed to assess the retrieval of anglicisms in Spanish. 
In other words, the sentences in BLAS thoroughly evaluate the ability of models at identifying a true positive (an anglicism) in different contexts and shapes. 
The results of the shared task show that even the best performing system systematically fails at retrieving anglicisms that contain words that also exist in Spanish (such as \textit{pie} or \textit{total red}), which proves that the retrieval of ambiguous anglicisms is still an unsolved task.

In addition, most participants added some sort of tailor-made heuristics to their systems, such as removing quotation marks, transforming the whole text to lowercase or consider foreign names in fist sentence position as anglicisms.
These heuristics tended to boost the scores obtained by the systems because BLAS was designed to assess recall over precision: in other words, sentences in BLAS were designed to be rich in anglicisms that are hard to identify (because of their shape, because of their context, etc.), but in exchange, BLAS does not thoroughly explore how models perform when sentences contain words that are likely to cause false positives errors (i.e., words that look like an anglicism but are not, such as odd-looking native words, native words written between quotation marks, foreign named entities, literal quotations, etc.).
In fact, even our poor-performing baselines produced good precision results on BLAS (see Table \ref{tab:baselines}), which illustrates that our test set is not challenging in terms of precision.
We wonder whether these heuristics that showed to be successful when dealing with BLAS could cause precision errors when dealing with sentences rich in potentially false positive examples.
For instance, one of the participating systems included a rule so that if a known anglicism appeared inside a longer sequence of text that was written between quotation marks, then the system was forced to span over the whole quoted text and return the whole quoted text as anglicism span.
This was a successful  approach when dealing with the examples contained in BLAS, but that strategy would have probably failed if the test set had contained a higher number of literal quotations or foreign named entities written between quotation marks: for instance, the word \textit{band} is likely to be a known anglicism (as in  \textit{big band}). If the named entity \textit{``Sgt. Pepper's Lonely Hearts Club Band''} had appeared in a sentence in the test set with quotation marks, that rule could make the system return the full entity as an anglicism.

Future work should thoroughly explore models' performance when dealing with sentences rich in false-positive examples where these simple heuristics fail.

\section{Conclusions}
In this paper we have introduced ADoBo 2025 shared task on automatic identification of English lexical borrowings in Spanish. 
We have presented BLAS, the dataset that was used for the test, and described the five systems that submitted results during the test phase.

Participating systems included several Transformer based solutions (such as XLM-R and BERT), various LLM with different prompting strategies (70B-Llama3.3, OpenAI models, etc.), a feed forward neural net fed with linguistic features and a lexicon-based rule system. Obtained results ranged from 17 to 99 on F1 score.
The best score was obtained by OpenAI o3 model prompted with extended guidelines and reminders, followed by the lexicon lookup system with contextual rules. These wide differences in performance showcase the impact that different approaches can have for this task.

In terms of errors, mistaking a named entity with a an anglicism or missing spans (either fully missing it or partially missing part of the span) were a common source of error. Ambiguous words (words that exist both as part of an anglicism or as a fully native word in Spanish, such as \textit{pie} or \textit{red}) were a challenge even for the best performing model.

\begin{table*}[t]
%\centering
\footnotesize
\centering
%\begin{sidewaystable*}
%\begin{adjustwidth}{-10cm}{0cm}
%\resizebox{600}{!}{
%\begin{tabular}{lcl}
\begin{tabular}{p{10.5cm}cl}
\toprule
\textbf{Example}  & \textbf{Prediction}	& \textbf{Error type}	
%\multicolumn{1}{c}{\textbf{Example}}  & \multicolumn{1}{c}{\textbf{Prediction}}	& \multicolumn{1}{c}{\textbf{Error type}}	
\\
\midrule
Un \underline{fatal error} ocurre cuando el programa intenta dividir por cero &
---  & Missing
\\%\hline

un \underline{fatal error} ocurre cuando el programa intenta dividir por cero &
--- & Missing
\\%\hline

UN \underline{FATAL ERROR} OCURRE CUANDO EL PROGRAMA INTENTA DIVIDIR POR CERO &
--- & Missing
\\ \\

Durante su carrera profesional, también ha sido socio y director general de Ikea, \underline{global director} de Apple y director de Comunicación de Basket Market. &
---  & Missing 
\\%\hline

durante su carrera profesional, también ha sido socio y director general de ikea, \underline{global director} de apple y director de comunicación de basket market. &
\textit{basket market} & Spurious, Missing
\\%\hline

DURANTE SU CARRERA PROFESIONAL, TAMBIÉN HA SIDO SOCIO Y DIRECTOR GENERAL DE IKEA, \underline{GLOBAL DIRECTOR} DE APPLE Y DIRECTOR DE COMUNICACIÓN DE BASKET MARKET. &
--- & Missing
\\%\hline

Durante su carrera profesional, también ha sido socio y director general de Ikea, \underline{GLOBAL DIRECTOR} de Apple y director de Comunicación de Basket Market. &
--- & Missing
\\\\

La reina Letizia ha escogido un conjunto \underline{total red} para la boda de los príncipes de de Holanda & \textit{red} & Overlap 
\\%\hline

la reina letizia ha escogido un conjunto \underline{total red} para la boda de los príncipes de de holanda &
\textit{red} & Overlap 
\\%\hline

La Reina Letizia Ha Escogido Un Conjunto \underline{Total Red} Para La Boda De Los Príncipes De De Holanda &
\textit{red} & Overlap 
\\%\hline

LA REINA LETIZIA HA ESCOGIDO UN CONJUNTO \underline{TOTAL RED} PARA LA BODA DE LOS PRÍNCIPES DE DE HOLANDA &
\textit{red} & Overlap 
\\\\

La actriz lució un \underline{look} \underline{total black} en el estreno de la película &
\textit{black} & Missing, Overlap 
\\%\hline

la actriz lució un \underline{look} \underline{total black} en el estreno de la película &
\textit{black} & Missing, Overlap 
\\\\

La Agencia Reivindica La Publicidad En Medios Clásicos Como La Radio Y La Televisión Y Desaconseja Fiarlo Todo A Campañas De ``\underline{Marketing}'' ``\underline{Online}''. &
\textit{marketing online} & Fused
\\%\hline

La agencia reivindica la publicidad en medios clásicos como la radio y la televisión y desaconseja fiarlo todo a campañas de \underline{Marketing} \underline{Online}. &
\textit{marketing online} & Fused
\\\\

``\underline{CASUAL LOOKS}'' CON BUFANDA Y GUANTES PARA TRIUNFAR ESTA TEMPORADA &
\textit{casual}, \textit{looks} & Split
\\%\hline

``\underline{Casual Looks}'' con bufanda y guantes para triunfar esta temporada &
\textit{casual}, \textit{looks} & Split
\\%\hline

\underline{casual looks} con bufanda y guantes para triunfar esta temporada &
\textit{casual}, \textit{looks} & Split
\\%\hline

\underline{Casual Looks} Con Bufanda Y Guantes Para Triunfar Esta Temporada &
\textit{looks} & Overlap 
\\%\hline

\underline{Casual Looks} con bufanda y guantes para triunfar esta temporada &
\textit{looks} & Overlap 
\\\\

tacones y vestidazos dejan hueco a \underline{casual looks} más alegres y festivos, donde hasta el chándal tiene protagonismo. &
\textit{casual}, \textit{looks} & Split
\\%\hline

Tacones Y Vestidazos Dejan Hueco A \underline{Casual Looks} Más Alegres Y Festivos, Donde Hasta El Chándal Tiene Protagonismo. &
\textit{casual}, \textit{looks} & Split
\\\\

\underline{Ugly Shoes} a todo color para para un verano fantástico &
\textit{ugly}, \textit{shoes} & Split
\\\\

Receta de \underline{Pie} de limón paso a paso y sin horno &
--- & Missing
\\\\

LOS DEFENSORES DEL \underline{NATURAL TIME} PROPONEN DISTRIBUIR EL CALENDARIO EN 13 MESES DE 28 DÍAS. &
\textit{time} & Overlap 
\\

\bottomrule
\end{tabular}%}
\caption{Error analysis of the errors produced by the best-performing system by \namecite{qilex}.}\label{tab:errors}
\end{table*}

\begin{table}[t]
\small
\centering
\begin{tabular}{lr}
\toprule
\multicolumn{1}{c}{\textbf{Team}} 
& \multicolumn{1}{c}{\textbf{F1-score}}	\\
\midrule
trockti  & 0.86 \\
trockti  & 0.83 \\
qilex  & 0.82 \\
shentzu  & 0.82 \\
trockti  & 0.81 \\
igorsterner  & 0.74 \\
igorsterner  & 0.74 \\
igorsterner  & 0.69 \\
igorsterner  & 0.52 \\
igorsterner  & 0.51 \\
trockti  & 0.23 \\
\bottomrule
\end{tabular}
\caption{F1 scores obtained on the development set per team.}
\label{tab:dev}
\end{table}

\begin{table}[t]
\small
\centering
\begin{tabular}{lr}
\toprule
\multicolumn{1}{c}{\textbf{Team}} 
& \multicolumn{1}{c}{\textbf{F1-score}}	\\
\midrule
qilex  & 0.99 \\ 
qilex  & 0.97 \\
shentzu  & 0.96 \\
shentzu  & 0.95 \\
mheredia  & 0.93 \\
trockti  & 0.92 \\
shentzu  & 0.90\\
mheredia  & 0.79 \\
hammond  & 0.75 \\
hammond  & 0.75 \\
hammond  & 0.74 \\
hammond  & 0.74 \\
hammond  & 0.72 \\
hammond  & 0.72 \\
hammond  & 0.72 \\
trockti  & 0.67 \\
mheredia  & 0.65 \\
hammond  & 0.65 \\
hammond  & 0.64 \\
hammond  & 0.64 \\
hammond  & 0.64 \\
hammond  & 0.64 \\
hammond  & 0.64 \\
hammond  & 0.64  \\
trockti  & 0.58 \\
trockti  & 0.47 \\
hammond  & 0.17 \\
\bottomrule
\end{tabular}
\caption{F1 scores obtained on the test set per team.}
\label{tab:test}
\end{table}

\bibliographystyle{fullname}
\bibliography{EjemploARTsepln}

\end{document}